\documentclass[10pt,twocolumn,letterpaper]{article}

\usepackage{cvpr}
\usepackage{times}
\usepackage{epsfig}
\usepackage{graphicx}
\usepackage{amsmath}
\usepackage{amssymb}
\usepackage{subfig}
\let \proof \relax

\usepackage{amsthm}
\usepackage{wrapfig}
\usepackage{algorithm}
\usepackage{algpseudocode}

\usepackage{mathrsfs}
\usepackage{booktabs}
\newcommand{\gt}{g_{\theta}}
\newcommand\fpfootnote[1]{%
  \begingroup
  \renewcommand\thefootnote{}\footnote{#1}%
  \addtocounter{footnote}{-1}%
  \endgroup
}

\usepackage{color}

\usepackage[pagebackref=true,breaklinks=true,letterpaper=true,colorlinks,bookmarks=false]{hyperref}

\newcommand{\ldiv}{\mathcal L_{ndiv}}
\newcommand{\lgen}{\mathcal L_{gen}}

\newcommand{\where}{\text{where}}
\ifx\proof\undefined
\newenvironment{proof}{\par\noindent{\bf Proof\ }}{\hfill\BlackBox\\[2mm]}
\fi

\ifx\theorem\undefined
\newtheorem{theorem}{Theorem}
\fi
\ifx\assumption\undefined
\newtheorem{assumption}[theorem]{Assumption}
\fi
\ifx\condition\undefined
\newtheorem{condition}{Condition}
\fi
\ifx\example\undefined
\newtheorem{example}{Example}
\fi
\ifx\definition\undefined
\newtheorem{definition}[theorem]{Definition}
\fi
\ifx\conjecture\undefined
\newtheorem{conjecture}[theorem]{Conjecture}
\fi
\ifx\fact\undefined
\newtheorem{fact}[theorem]{Fact}
\fi
\ifx\claim\undefined
\newtheorem{claim}[theorem]{Claim}
\fi

\cvprfinalcopy 

\def\cvprPaperID{4658} 

\ifcvprfinal\fi
\begin{document}

\title{Normalized Diversification}

\author{
Shaohui Liu$^{1*}$ \quad  Xiao Zhang$^{2*}$ \quad Jianqiao Wangni\textsuperscript{2} \quad Jianbo Shi\textsuperscript{2} \\ 
$^1$Tsinghua University \qquad $^2$University of Pennsylvania \\
{\tt\small b1ueber2y@gmail.com, \{zhang7, wnjq,  jshi\}@seas.upenn.edu}
}

\maketitle
\thispagestyle{empty}

\begin{abstract}
\fpfootnote{$\ast$ indicates equal contribution}


Generating diverse yet specific data is the goal of the generative adversarial network (GAN), but it suffers from the problem of mode collapse. We introduce the concept of normalized diversity which force the model to preserve the normalized pairwise distance between the sparse samples from a latent parametric distribution and their corresponding high-dimensional outputs. The normalized diversification aims to unfold the manifold of unknown topology and non-uniform distribution, which leads to safe interpolation between valid latent variables. By alternating the maximization over the pairwise distance and updating the total distance (normalizer), we encourage the model to actively explore in the high-dimensional output space. We demonstrate that by combining the normalized diversity loss and the adversarial loss, we generate diverse data without suffering from mode collapsing. Experimental results show that our method achieves consistent improvement on unsupervised image generation, conditional image generation and hand pose estimation over strong baselines. 

\end{abstract}
\vspace{-10pt}

\section{Introduction}

Diversity is an important concept in many areas, e.g. portfolio analysis \cite{solnik1974not}, ecological science \cite{magurran1988diversity} and recommendation system \cite{ziegler2005improving}. This concept is also crucial to generative models which have wide applications in machine learning and computer vision. Several representative examples include Variational Autoencoder (VAE) \cite{kingma2013auto} and Generative Adversarial Network (GAN) \cite{goodfellow2014generative}, which are capable of modeling complicated data.  One ideal principle shared by all generative models, simple or complex, is quite similar, that the generated data should be diverse. Otherwise, the model may have a so-called problem of mode collapse, where all generated outputs are highly similar. This problem is more common in GAN \cite{goodfellow2014generative} since the objective function is mostly about the validity of generated samples but not the diversity of them. 

\begin{figure}
    \centering
    \includegraphics[width=0.5\textwidth]{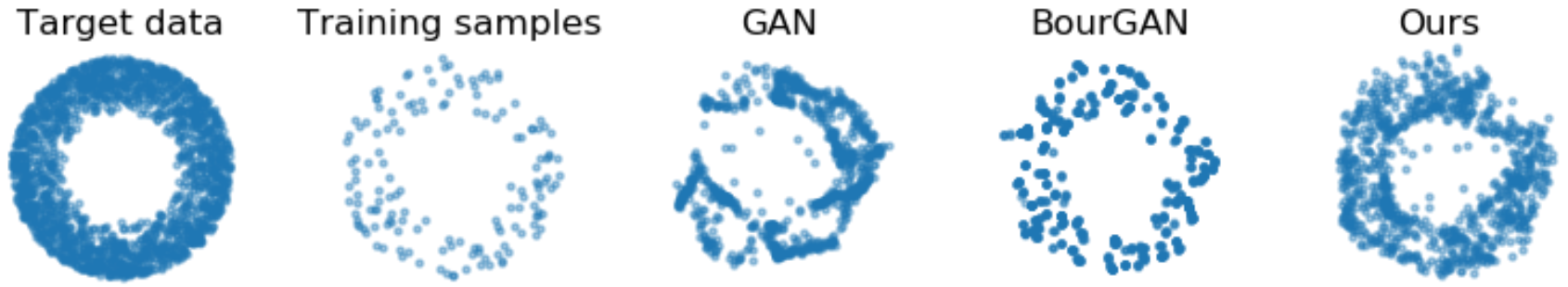}
    \caption{Comparison of generative models' capability to learn from sparse samples with unknown topology (a donut shape). 
    Generated samples from GAN \cite{goodfellow2014generative}, BourGAN \cite{xiao2018bourgan} and ours are illustrated. 
    \textbf{GAN} \cite{goodfellow2014generative} suffers from mode collapse.  
    \textbf{BourGAN} \cite{xiao2018bourgan} concentrates tightly around the training data points. \textbf{Ours} generates dense coverage around the training samples with limited outliers.}
    \label{fig::donut-demo}
    \vspace{-10pt}
\end{figure}

Our goal is to learn a generative model for high-dimensional image data which are non-uniformly distributed over a space with unknown topology, only using very sparse samples. These conditions stated above, even one of them being removed, may result in a much easier problem to solve. 
We formulate the solution as learning a mapping from a low-dimensional variable in latent space $\mathcal Z$ to an image defined in output space $\mathcal Y$. However, we face a dilemma that, in order to fit the complicated topology of output data, we need the learned mapping to be complex and highly expressive, which actually requires dense samples from real-world. This problem might be alleviated if the topology is simpler since a parametrized function of smaller VC dimension might be able to fit. This situation however is most likely not the case if we need to deal with arbitrary images. 

%

\begin{figure*}
\centering

\renewcommand{\arraystretch}{0.2}
\setlength\tabcolsep{1.8pt}
\begin{tabular}{|c|cccccccc}
\cline{1-1}
 & & norm. $\gt(z)$ & & norm. $\gt(z)$ & & norm. $\gt(z)$ & & $\gt(z)$ 
\\
$\mathcal{Z}$ & Target $\mathcal{Y}$ & pairwise dist. & iter. 1000 & pairwise dist. & iter. 4000 & pairwise dist. & iter. 8000 & iter. 8000
\\
{\includegraphics[width=1.72cm, height=1.72cm, trim={15 15 15 15}, clip]{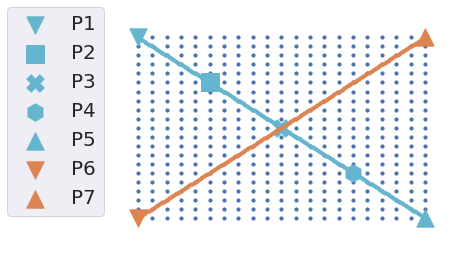}} &
{\includegraphics[width=1.72cm, height=1.72cm, trim = {35 35 35 35}, clip]{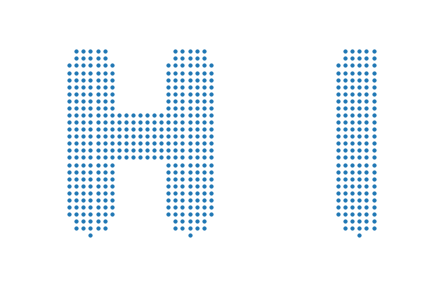}} &
{\includegraphics[width=1.72cm, height=1.72cm]{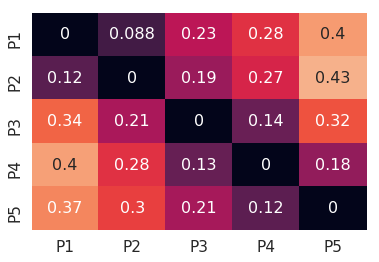}} &
{\includegraphics[width=1.72cm, height=1.72cm, trim={20 20 20 20}, clip]{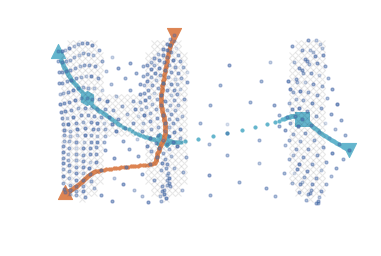}} &
{\includegraphics[width=1.72cm, height=1.72cm]{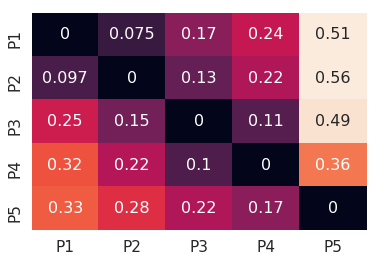}} &
{\includegraphics[width=1.72cm, height=1.72cm, trim={20 20 20 20}, clip]{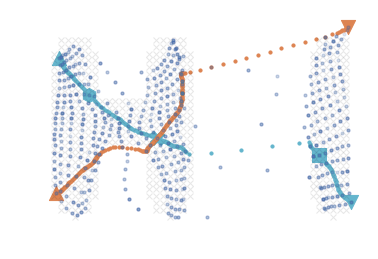}} &
{\includegraphics[width=1.72cm, height=1.72cm]{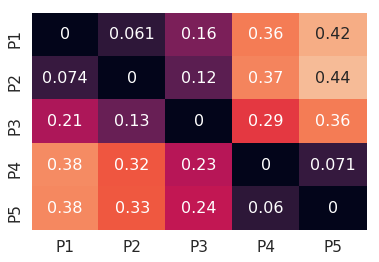}} &
{\includegraphics[width=1.72cm, height=1.72cm, trim={20 20 20 20}, clip]{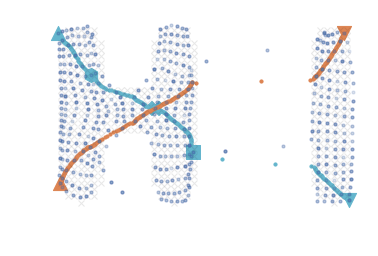}} &
{\includegraphics[width=1.72cm, height=1.72cm, trim={35 35 35 35}, clip]{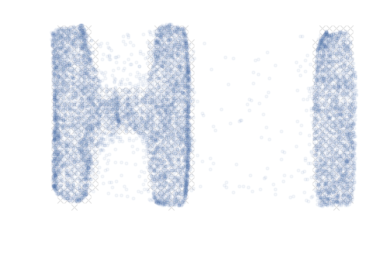}}
\\
norm. $z$ & & norm. $\gt(z)$ & & norm. $\gt(z)$ & & norm. $\gt(z)$ & & $\gt(z)$
\\
pairwise dist. & Target $\mathcal{Y}$ & pairwise dist. & iter. 100 & pairwise dist. & iter. 200 & pairwise dist. & iter. 1000 & iter. 1000
\\
{\includegraphics[width=1.72cm, height=1.72cm]{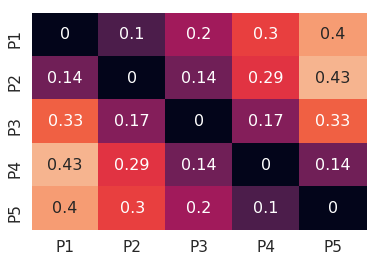}} &
{\includegraphics[width=1.72cm, height=1.72cm, trim={30 30 30 30}, clip]{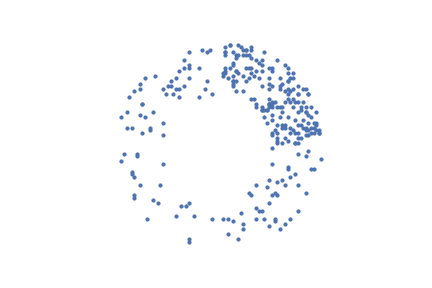}} &
{\includegraphics[width=1.72cm, height=1.72cm]{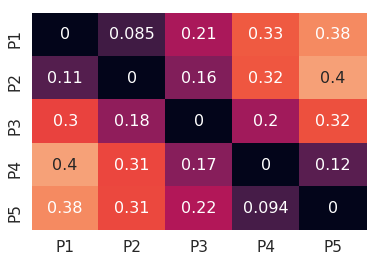}} &
{\includegraphics[width=1.72cm, height=1.72cm, trim={20 20 20 20}, clip]{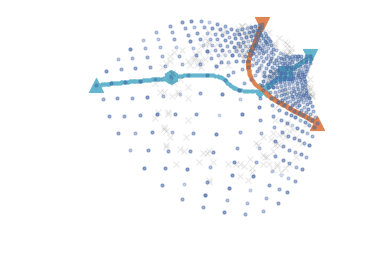}} &
{\includegraphics[width=1.72cm, height=1.72cm]{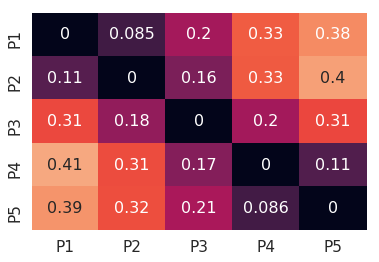}} &
{\includegraphics[width=1.72cm, height=1.72cm, trim={20 20 20 20}, clip]{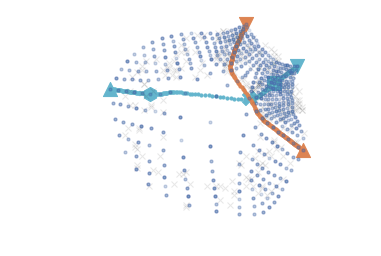}} &
{\includegraphics[width=1.72cm, height=1.72cm]{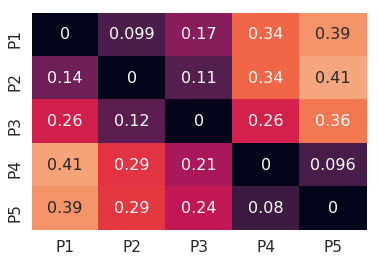}} &
{\includegraphics[width=1.72cm, height=1.72cm, trim={20 20 20 20}, clip]{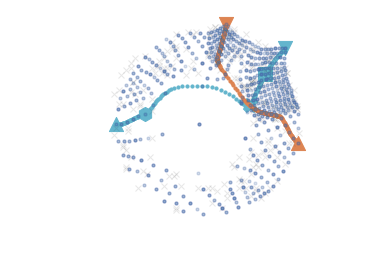}} &
{\includegraphics[width=1.72cm, height=1.72cm, trim={25 25 25 25}, clip]{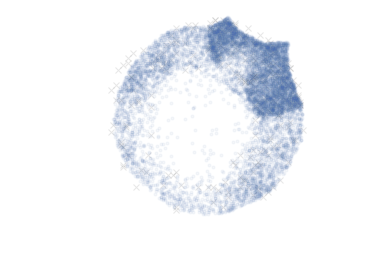}}
\\
\cline{1-1}
\end{tabular}

\caption{Illustration on the training procedure with proposed normalized diversification on highly irregular  
topology (top row) and non-uniform data density (bottom row). The generative model can effectively learn from sparse data by constructing a mapping from the latent space $\mathcal Z$ to the target distribution by minimizing normalized diversity loss. \textbf{Top Left:} The latent variable $z \in \mathcal{Z}$ in 2D space is sampled from an uniform distribution. 5 points (P1-P5, colored blue) along the diagonal are used for illustration. 
\textbf{Bottom Left:} The normalized pairwise distance matrix on P1-P5. 
\textbf{From Left to Right:} We show qualitative results on two synthetic cases: `HI' and `Ring'. 
We visualize the mapping from the latent space to the output space for several iterations together with $D^Y \in \mathbb R^{5 \times 5}$ and 
we illustrate safe interpolation on the diagonal of the latent space onto the output space. 
\textbf{Right Most Column:} We generate dense samples from the learned model, to illustrate the diversity and the consistency w.r.t.  
 the ground-truth distribution shown in the second column.
}
\label{fig::maindemo}
\vspace{-10pt}
\end{figure*}

We start from an idea that is orthogonal to previous research: that we aim to expand as well as unfold the manifold of generated images, actively and safely. We want to model the data with  complex topology in output space using a parametrized mapping from  a latent space with simple topology. To learn a model that is specific to training images and also generalizable, we wish to achieve that the  \textit{interpolation} and \textit{extrapolation} on a pair of latent variables should generate valid samples in output space. 
This requires the generative model, globally, to generate diverse and valid images via exploration, which we refer to as safe extrapolation, and locally, to preserve the neighbouring relation so that interpolation is also safe. 

In this paper, we propose \textit{normalized diversification} following the motivation above. This is achieved by combining the adversarial loss of GAN and a novel normalized diversity loss. This diversity loss encourages the learned mapping to preserve the normalized pairwise distance between every pair of inputs $z,z' \in \mathcal Z$ and their corresponding outputs $\gt(z),\gt(z') \in \mathcal Y $, where $\gt$ is a parametrized function. This utilizes the same insight as manifold unfolding \cite{weinberger2006distance}. 
During training, we also fix the normalization term while maximizing the pairwise distance, which also encourages the model to visit outer modes i.e. to extrapolate. 


To illustrate the concept, we sample sparse points from a donut shape as training data for GAN \cite{goodfellow2014generative}, BourGAN \cite{xiao2018bourgan} and our method, shown in Figure \ref{fig::donut-demo}. 
After the model being trained, we generate 5k new samples from the learned distribution for visualization. Our method achieves safe interpolation, fills in the gap in sparse samples and generates dense coverage while GAN \cite{goodfellow2014generative} and BourGAN \cite{xiao2018bourgan} fails to generalize. 


Our paper is organized as follows: Section \ref{sec::motivation} presents more motivations towards a better understanding of normalized diversification. Section \ref{sec::method} describes how we apply normalized diversification onto multiple vision applications by employing different metric spaces on the outputs. Finally in Section \ref{sec::exp}, we show promising experimental results on multiple tasks to demonstrate the effectiveness.

\section{Related Work}
\label{sec::related_work}
Early attempts for addressing the problem of mode collapse include maximum mean discrepancy \cite{ dziugaite2015training}, boundary equilibrium \cite{berthelot2017began} and training multiple discriminators \cite{durugkar2016generative}. Later, some other works \cite{arjovsky2017wasserstein, che2015mode, metz2016unrolled} modified the objective function. The usage of the distance matrix was implicitly introduced as a discrimination score in \cite{NIPS2016_6125} or covariance in \cite{karras2017progressive}. Recently, several novel methods via either statistical or structural information are proposed in \cite{srivastava2017veegan, lin2017pacgan, xiao2018bourgan}.

Most problems in computer vision are fundamentally ill-posed that they have multiple or infinite number of solutions. 
To obtain a better encoder with diverse generated data, there are a variety of ideas making full use of the VAE \cite{kingma2013auto}, GAN \cite{goodfellow2014generative} and its conditional version to develop better models 
\cite{bao2017cvae, isola2017image, larsen2016autoencoding, reed2016generative, spurr2018cross, zhu2017toward}.  Huang et al. \cite{huang2018multimodal} proposed to learn a disentangled representation, Ye et al. \cite{ye2017occlusion} used a parametrized GMM, other works like \cite{bansal2017pixelnn,chen2017photographic,fan2017point,firman2018diversenet} selectively back-propagated the gradients of multiple samples. The {normalized diversification}, defined with pairwise terms on the mapping function itself, appears orthogonal to most of the existing methods.


\section{Method}
\label{sec::motivation}

We consider a generative model $\gt$ that generates an image $y \in \mathcal Y$ based on a latent variable $z$. The target of training the model is to fit an unknown target distribution $p_{data}(y)$ based on limited samples. In this paper, we consider two different kinds of 
implicit generative models that solve different problems but intrinsically share the same spirit of diversification. 
\begin{itemize}
	\item \textit{Unsupervised generative model.} This model is used for tasks that do not depend on auxiliary information, e.g. image synthesis without supervision. Existing methods such as GAN \cite{goodfellow2014generative} and VAE \cite{kingma2013auto} use a latent variable $z \in \mathcal{Z}$ that follows a parametric distribution $p(z)$ in latent space and learn the mapping $\gt:\mathcal{Z} \rightarrow \mathcal{Y}$ to fit the target $y \in \mathcal{Y}$.
	\item \textit{Conditional generative model. } This model utilizes additional information to generate more specific outputs, e.g. text-to-image (image-to-image) translation, pose estimation, future prediction. Related works include a variety of conditional generative models \cite{bao2017cvae, isola2017image, larsen2016autoencoding, reed2016generative, zhu2017toward}. They also use a predefined latent space $\mathcal{Z}$ and aim to fit the joint distribution $(\mathcal{X}, \mathcal{Y})$ for the input domain $\mathcal{X}$ and output domain $\mathcal{Y}$ (for convenience we also use $\mathcal{Y}$ here). Specifically, an encoder $E:\mathcal{X}\rightarrow \mathcal{C}$ is used to get the latent code $c\in \mathcal{C}$. Then, either addition (VAE) or concatenation (CGAN) is employed to combine $c$ and $z$ for training generator $\gt:\mathcal{C}\times \mathcal{Z}\rightarrow \mathcal{Y}$.
\end{itemize}
The problem of mode collapse is frequently encountered especially in generative models like GAN \cite{goodfellow2014generative}, that the model generates highly similar data but satisfies the criterion for training. 
Our motivation for normalized diversification is to encourage $\gt$ to generate data with enough diversity, so the model visits most of the important modes. In the meantime, we wish  $\gt$ to be well-conditioned around the visited modes, so we could infer latent variables from some valid image samples, and ensure safe interpolation and extrapolation between these latent variables to generate meaningful images. 
We could address the problem of mode collapse by diversifying the outputs, which is similar to enlarging the variance parameter of a Gaussian distribution, but a hard problem is how to measure the diversity of real-world images analytically and properly, as the variance of  a distribution might be too universal for general tasks to be specific to our problem. 

We start with an intuition to prevent the pairwise distance $d(\cdot,\cdot)$ between two generated points $A$ and $B$ from being too close. Note that if the outputs are linearly scaled, so will the pairwise distance $d(A,B)$: the samples seem to diversify from each other, without actually solving the intrinsic problem of mode collapse. 
Thus, we measure whether the mapping preserves the normalized pairwise distance between inputs $\|z-z'\|$ and outputs $\|\gt(z)-\gt(z')\|$.

Since we do not have access to infinite amount of data,  we sample a limited amount of data from a well-defined parametric distribution $p(z)$, and try to visit more mode via \textit{diversification}. We measure the diversity of generated samples by the pairwise distance in $\mathcal Y$ through the parametrized mapping $\gt$. The objective function is simplified as a finite-sum form on $N$ samples $\{z_i\}_{i=1}^N$, along with corresponding images $\{y_i| y_i=\gt(z_i)\}_{i=1}^N$. We denote two metric spaces $M_Z=(\mathcal{Z}, d_Z)$ and $M_Y=(\mathcal{Y}, d_Y)$. We use two additional functions $h_Y$ and $h_Z$ for some task-specific usage, and then define the metric as composite functions upon Euclidean distance, as follows,
\begin{align}
d_Z(z_i, z_j)=\|h_Z(z_i) - h_Z(z_j)\|_2
\label{eq::emd_z} \\
d_Y(y_i, y_j)=\|h_Y(y_i)-h_Y(y_j)\|_2
\label{eq::emd_y}
\end{align}
To explain with a concrete application in diverse video synthesis that we try to generate a realistic video from a moving vehicle based on a sequence of segmentation mask, here $d_Y$ may be used for extracting deep features using an off-the-shelf network like the perceptual loss \cite{johnson2016perceptual}. 
We also define $D^Z, D^Y\in \mathbb{R}^{N \times N}$ to be the normalized pairwise distance matrix of $\{z_i\}_{i=1}^N$ and $\{y_i\}_{i=1}^N$ respectively in Eq.(\ref{eq::ndistance_z}). Each element in the matrix is defined as
\begin{eqnarray}
D^Z_{ij} = \frac{d_Z(z_i, z_j)}{\sum_{\substack{j}} d_Z(z_i, z_j)}
\label{eq::ndistance_z}, \quad
D^Y_{ij}=\frac{d_Y(y_i, y_j)}{\sum_{\substack{j}} d_Y(y_i, y_j)}
\label{eq::ndistance_y},
\end{eqnarray}
for $\forall i, j \in[N]$. The normalized diversity loss can be formulated as in Eq.(\ref{eq::div}), where $\alpha \approx 1 $ is a slack factor of this diversity loss.
{
\begin{equation}
\ldiv(\theta, p)= \frac{1}{N^2-N} \sum_{i=1}^N \sum_{j \neq i}^N \max(\alpha D^Z_{ij}-D^Y_{ij},0).
\label{eq::div}
\end{equation}
}
To enforce the extrapolation, we treat the normalizer in Eq.(\ref{eq::ndistance_z}) as a constant when back-propagating the gradient to generator. As a result minimizing Eq.(\ref{eq::div}) would also force the model to actively expand in the high-dimensional output space, we refer to Algorithm \ref{alg::pipeline_cross} and Section \ref{interp_extrap} for more details. 

\begin{figure}
    \centering
    \includegraphics[width=0.5\textwidth]{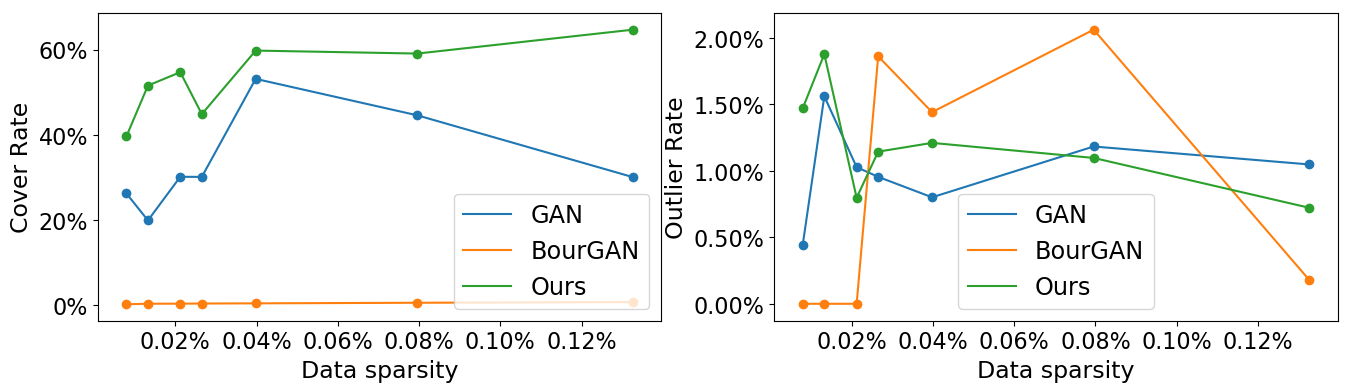}
    \caption{Quantitative comparison between normalized diversity loss and BourGAN loss \cite{xiao2018bourgan} on the learned distribution in Fig. \ref{fig::donut-demo}. We discretized the donut region into uniform mesh grids and measured ``cover rate'' (the percentage of grids which have generated samples in them). We also measured ``outlier rate'': the ratio of samples outside the donut. ``Data sparsity'' measures the cover rate of the training samples over the donut. Our method improve the cover rate over GAN \cite{goodfellow2014generative} and BourGAN \cite{xiao2018bourgan} while maintaining comparable outlier rate.}
    \label{fig::donut-curve}
\end{figure}

For the purpose of unfolding the manifold, we could focus on the expansion of densely connected pairs of $ D^Z_{ij}>D^Y_{ij}$  or on the contraction of loosely connected pairs of $D^Z_{ij}<D^Y_{ij} $.
Only minimizing $\max(\alpha D^Z_{ij}-D^Y_{ij},0)$ would encourage active extrapolation when we hold the normalizer constant.


Combining the diversity loss, the objective function over $\theta$ can be written in a compact form as
\begin{align}
\min_{\theta} \mathcal L(\theta, p)=\lgen (\theta,p) +\ldiv(\theta, p)
\label{eq::total_loss}
\end{align}
where $\lgen(\theta,p)$ is the original objective function for learning the generator $\gt$. In the conditional model, the objective function also depends on $x$ which we omitted here. 

\subsection{Interpolation and Extrapolation}\label{interp_extrap}

A fundamental motivation behind the normalized diversification is to achieve that \textit{interpolation} and \textit{extrapolation} on a pair of latent variables should generate valid samples in output space. This pursues a trade-off between pairwise diversity and validity.  This motivation aligns with the concept of \textit{local isometry} \cite{weinberger2006unsupervised},  which has a nice property that the near neighbors in $\mathcal Z$, are encoded to $\mathcal Y$ using a rotation plus a translation. Local isometry requires the output manifold to be smooth and without invalid `holes' within the neighbor region of a valid point, and the interpolation in $\mathcal Z$  generates valid points in $\mathcal Y$ through a parametrized mapping $\gt$ where $\theta$ is the learned parameter. Differently, Variational Autoencoder (VAE) \cite{kingma2013auto} mostly cares about local perturbation, or pointwise diversity. 

%
Normalized diversification is orthogonal to previous research as it aims to unfold the manifold of generated images \cite{weinberger2006distance}. 
\textbf{Interpolation.} By pushing apart the pairwise distance of the sample points, we prevent the 'short-cuts' that links samples through the exterior space.  As shown in Figure \ref{fig::donut-demo}, given a set of points, our goal is to discover the underlying parametrization so we can densely generate new valid samples in the interior (on the donut),  without crossing over to the exterior (the donut hole). This leads to safe interpolation.
\textbf{Extrapolation.} For active exploration of output space $\mathcal Y$ with the current model, in each iteration, we first calculate the normalizer of the pairwise distance matrix, then use the gradient back-propagated from the $d^Y_{ij}$ to force expansion, 
after which we update the normalizer. With these alternating steps, we ensure the stability of exploration.  
We illustrate the evolution of the training procedure on two synthetic 2D distributions 
in Figure \ref{fig::maindemo}


\subsection{Understanding from Geometric Perspective}
Simply trying to enlarge the pairwise distance between samples in $\mathcal Y$ can explore the unobserved space, but a crucial problem is how to make sure the interpolated points are still valid. From a geometric perspective, imagine that $A, B, C$ are on a curvy $1D$ line in $2D$ space, the transitive distance between them, $d(A,B)+d(B,C)$ on the 1D curve is much longer than 2D distance $d(A,C)$, which violates the triangle inequality. If we make an interpolation point $D$ in the inner part of the line $(A,C)$, although the direct 2D distance between $d(A,C)$ could be very small, the line between them might lie in the part of unreasonable space out of the manifold. However, by pushing $(A,C)$ as far as possible away, we `discover' the true 1D distance. 

This insight is different from existing approaches, e.g. BourGAN \cite{xiao2018bourgan}, which aims at matching the pairwise distance between output space and latent space which can preserve the data modality but hinder its generalization ability. Our method, besides the mode preserving ability, can also actively expand and unfold the manifold which enables safe interpolation on the latent space where the generated samples will neither lie outside the valid region nor overfit the existing data (See Figure \ref{fig::donut-demo} and \ref{fig::donut-curve}).


\subsection{Understanding via Simplified Models}
To understand the normalized diversification, we start from another perspective by using simple functions to illustrate the functionality of the regularization term. We assume the generator to be a simple linear model as $\gt(z)=\theta^T z$, where $\theta$ is the matrix that characterizes the linear transformation. A simple calculation induces that the diversity regularization thus encourages $\|\theta\|_{\star}$ (or $\|\theta\|_{2}$) being sufficiently large. However, this alone does not prevent some degenerated cases. Suppose that the generator is constructed in the following way 
\begin{align*}
\gt(z)= \theta^T z, \quad \theta=U^T diag[\beta ,0,0,\cdots,0] V\in \mathbb R^{K \times D},
\end{align*}
where $K$ and $D$ are the dimension of the latent variable $z$ and generated data respectively, and $U,V$ are the matrices consisting of all singular-vectors. When $\beta$ is sufficiently large, this model seems to be strongly diversified, however, is actually not a reasonable model as it measures the difference of two vectors along only one direction, i.e. the singular-vector corresponding to the singular value $\beta$. Normalization over the diversity helps to prevent these kinds of degenerated cases, as the normalized distance does not scale with $\beta$. The normalization also helps to adapt to a conditional setting that some related works e.g. BourGAN \cite{xiao2018bourgan} might fail to adapt, based on an input variable $x$ and a latent variable $z$, especially with large variations w.r.t. $x$. 
Imagine a simple example where $y=\gt(x,z) = (x+z)^3$, and $x \in [1,10]$. The upper bound of Lipschitz constant could be larger than $3 {\max(x)}^2=300$ while the lower bound should be lower than $3 {\min(x)}^2=3$. However, this value is meaningless for the most part of the whole domain $[1,10]$. 
Normalizing the distance could fix it to a reasonable range.  
Another side problem is on how to keep the semi-definiteness of the distance metric, to concord with metric learning research \cite{davis2007information, weinberger2006distance, weinberger2006unsupervised, xing2003distance}. Under this property, a pair of samples $(z,z')$ deviating from each other in whichever direction, should increase the pairwise distance in the output space. This property should also be enforced by diversification, although may not explicitly. This formulation of taking nonnegative part using $\max(\cdot, 0)$  encourages the learned metric in $\mathcal Y$ to be positive semidefinite (PSD). To understand with a counter-example, 
we consider a metric defined as
\begin{align*}
    d_Y(y,y')=(y-y')^T \Phi (y-y'), \quad \where \quad \Phi  \in \mathbb R^{D \times D}. 
\end{align*}
We assume the eigen-decomposition of $\Phi$ to be 
\begin{align}
    \Phi=U^T diag[-\beta ,100\beta ,0,0,\cdots,0] U
\end{align}
where $\beta>0$ (eigen-decomposition), and $\Phi$ is not PSD. 
Suppose there is a pair of latent variables $(y,y')$ that $y-y'= \sum_d U_d /D$, where summation is taken over all eigenvectors. This pair of data incurs a variation of $99 \beta$ of $\ldiv$ in Eq.(\ref{eq::div}) if the nonnegative operator $\max(\cdot,0)$ is not used, which leads to the objective function being minimized if $\beta$ positively scales, but clearly, the metric is not reasonable as it is not PSD. However, if the operator is applied, the same case incurs a variation of $\beta$ of $\ldiv$ from a deviation in the subspace corresponding to the negative eigenvalue $-\beta$, as the $\max(\cdot,0)$ operator strongly contrast those directions like $U_1$ that wrongly contribute to $\ldiv$, and ignore other subspace even if they correspond to strongly positive eigenvalues.
\begin{figure}[tb]
	\centering
	\includegraphics[width=240pt]{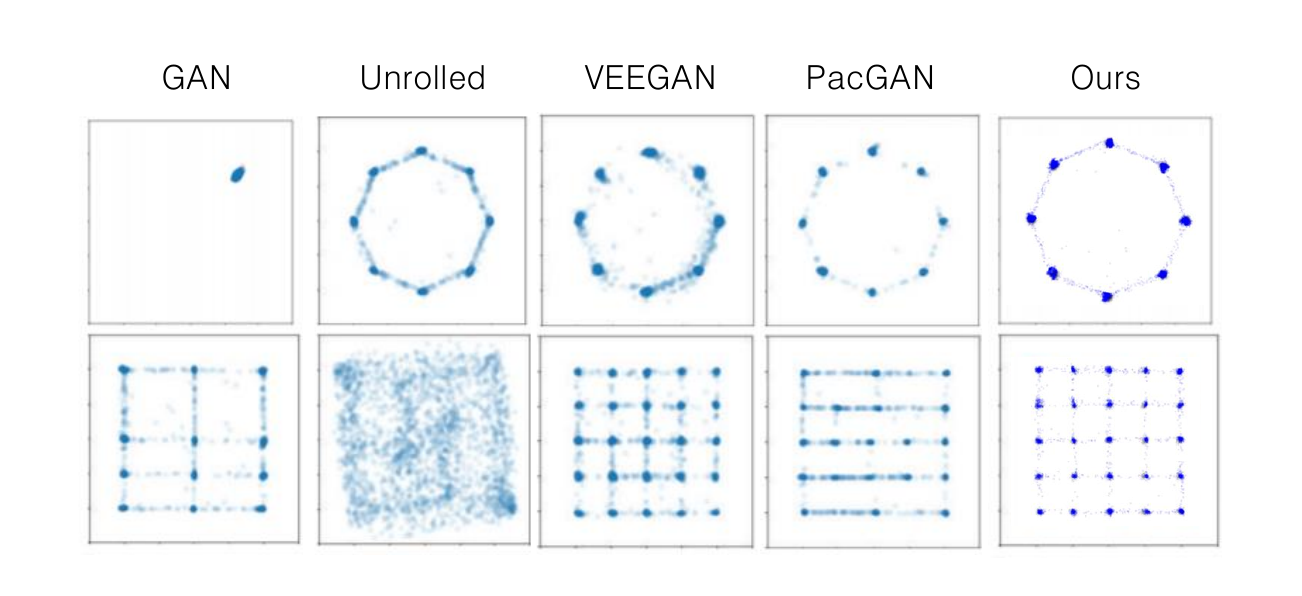}
	\caption{Qualitative results on `2D Gaussian ring' and `2D Gaussian grid'. For baseline methods, We directly used the visualization results in \cite{xiao2018bourgan}. 
    }
	\label{fig::5x5}
	\vspace{-10pt}
\end{figure}

\section{Real-World Applicability}

\label{sec::method}

In this section, we present the real-world applicability of normalized diversification for different vision applications. 
For different tasks, we need a validity checking function $F$ of the generated samples to validate that the data satisfies some domain-specific constraints. For example, for the unsupervised setting, $F$ can be interpreted as the trained discriminator in GAN \cite{goodfellow2014generative}. 
Our training pipeline is summarized in Algorithm \ref{alg::pipeline_cross}. For a given latent distribution $p(z)$ and the target distribution $p_{data}$, we sample a finite set of latent samples $\{z_i\}_{i=1}^n$ from $p(z)$ and $\{y_i\}_{i=1}^n$ from $p_{data}$. We then compute normalized pairwise distance matrix over $\{z_i\}_{i=1}^n$ and the generated samples $\{\gt(z_i)\}_{i=1}^n$ with the functions $d_Z$ and $d_Y$. We update the normalizer and block its gradients as previously discussed. Finally, we compute the total loss $\mathcal{L}$ in Eq.(\ref{eq::total_loss}) and update the model with the back-propagated gradients for each step.


\begin{algorithm}[tb]
	\caption{
	Training generative model with normalized diversity.
    }
	\label{alg::pipeline_cross}
	\begin{algorithmic}[1]    
	    \State \textbf{Given}: {Latent distribution $p(z)$, Target distribution $p_{data}$}
        \State \textbf{Given}: {Current generator $g_{\theta}$, Distance function $d_Z$, $d_Y$}
        \Function{NORMDIST} {$\{q_j\}_{i=1}^N, d$}:
        \For{$i \gets 1$ to $N$}
            \State {$S_i = \sum_{j=1}^N d(q_i,q_j)$ } 
            \State $S_i \leftarrow S_i.value$ (treated as constant value with back-propagated gradients blocked)
            \State {$D_{ij} = d(q_i,q_j)/S_i$, $j \in [N]$}
        \EndFor
        \State \Return $D$
        \EndFunction
        \While {not converged}
        \State {$\{z_i\}_{i=1}^N \sim p(z)$}
        \State {$\{y_i\}_{i=1}^N \sim p_{data}$}
        \State {$D^Z = $NORMDIST($\{z_i\}_{i=1}^N,d_Z$)}
	    \State {$D^Y = $NORMDIST($\{\gt(z_i)\}_{i=1}^N,d_Y$)}
	    \State (Optional for GAN) Update the discriminator $F$.
	    \State {Compute total loss $\mathcal{L}$ by Eq. \ref{eq::total_loss}}
	    \State {Update model parameter $\theta = \theta - \eta \partial \mathcal{L} / \partial \theta $}
	    \EndWhile
	\end{algorithmic}
\end{algorithm}


\subsection{Applications}
\label{sec::method_text2image}
The proposed normalized diversity term is general and can be applied to many vision applications including image generation, text-to-image (image-to-image) translation and hand pose estimation, etc. We specify the conditional input domain $\mathcal{X}$, the output domain $\mathcal{Y}$ and the function $h_Y$ for each application to compute normalized diversity loss in Eq.(\ref{eq::div}). We use a predefined $\mathcal{Z}$ with uniform distribution and $h_z(z)=z$ for all tasks.
\begin{itemize}
    \item \textbf{Image generation.} There is no conditional input for this unsupervised generative model setting. $\mathcal{Y}$ is the output image domain. The normalized pairwise distance on $\mathcal{Y}$ can be computed either with $\ell_1$ or $\ell_2$ distance or employing deep metrics such as using the GAN \cite{goodfellow2014generative} discriminator as the function $h_Y$.
    \item \textbf{Conditional image generation.} The conditional input domain $\mathcal{X}$ can be the input text and input image for text-to-image and image-to-image translation respectively. $\mathcal{Y}$ is the output image domain. The computation of normalized pairwise distance on $\mathcal{Y}$ is similar to that in image generation.
    \item \textbf{Hand pose estimation.} The conditional input domain $\mathcal{X}$ is the input RGB hand image. $\mathcal{Y} \in \mathbb{R}^{3K}$ is the output 3D pose, where $K$ denotes the number of the joints. For the function $h_Y$, we employ the visibility mask $V \in \{0,1\}^{3K}$ to gate each joint of the output 3D pose inversely, encouraging diversity on the occluded joints, i.e. $h_Y(y)=(1-V)\circ y$.
\end{itemize}

Note that \cite{xiao2018bourgan} cannot fit in the conditional setting because the number of possible conditional inputs, e.g. input text, RGB hand image are infinite, making the pairwise information for each conditional input impossible to be precomputed. Specifically for hand pose estimation, we develop a pseudo-renderer by using morphological clues to get a visibility mask.

The application of normalized diversity can fit in arbitrary checking function $F$ for different tasks. When compared to baseline adversarial methods, we use adversarial loss for image generation either unsupervised or conditional. For hand pose estimation, we choose to combine the $l_2$ distance on the visible joints and the joint adversarial discriminator of $(x_i, \hat{y}_i)$ (image-pose GAN), where $\hat{y}_i$ denotes the 2D projection of the output 3D pose $y_i$. 
\begin{align}
F(x,y)=F_{vis}(x,y)+F_{gan}(x,y)
\end{align} 
The two checking functions 
are formulated as follows 
\begin{align}
&	F_{vis}(x, y)=\|V\circ (y-y_r)\|_2^2\\
\label{eq::check_vis}
F_{gan}(x, y)&=\mathbb{E}_{x,\hat{y}_r \sim p_{data}}[\log D(x, \hat{y}_r)] \\&+ \mathbb{E}_{x \sim p_{data}, z\sim p(z)}[\log(1- D(x, \hat{y}))]\nonumber
\label{eq::check_gan}
\end{align}
With the image-pose GAN, the system has the capability to learn much subtle relationship between $x$ and $y$. By constraining only on the visible joints, our system gets less noisy gradients and can learn better pose estimation along with better image features.

\begin{table}
	\centering
	\caption{Results on the synthetic dataset. We followed the experimental setting in \cite{xiao2018bourgan}.}
	\begin{tabular}{*{5}{c}}
		\toprule 
		& \multicolumn{2}{c}{\textbf{2D Ring}} & \multicolumn{2}{c}{\textbf{2D Grid}}\\
		\cmidrule(lr){2-3}\cmidrule(lr){4-5}
		Method & \#modes & fail (\%) & \#modes & fail (\%) \\
		\midrule
		\midrule
		GAN \cite{goodfellow2014generative} & 1.0 & 0.1 & 17.7 & 17.7 \\  
		Unrolled \cite{metz2016unrolled} & 7.6 & 12.0 & 14.9 & 95.1 \\
		\textbf{Ours} & \textbf{8.0} & 10.3 & \textbf{25.0} & \textbf{11.1} \\
		\midrule
		\midrule
		VEEGAN \cite{srivastava2017veegan} & 8.0 & 13.2 & 24.4 & 22.8 \\
		PacGAN \cite{lin2017pacgan} & 7.8 & 1.8 & 24.3 & 20.5 \\
		BourGAN \cite{xiao2018bourgan} & 8.0 & 0.1 & 25.0 & 4.1 \\
		\bottomrule
	\end{tabular}
	\label{tab::5x5}
	\vspace{-10pt}
\end{table}

\section{Experiments}
\label{sec::exp}
\begin{table}[tb]
	\centering
	\caption{FID Results \cite{NIPS2017_7240} on Image Synthesis. ``SN" and ``GP" denote spectral normalization \cite{miyato2018spectral} and gradient penalty \cite{gulrajani2017improved} respectively. Our method achieved consistent improvement over various GANs.}
	\label{tab::gan_multiple}
	\begin{tabular}{*{5}{c}}
		\toprule
		& \multicolumn{2}{c}{CIFAR-10} & \multicolumn{2}{c}{CelebA}\\
		\cmidrule(lr){2-3}\cmidrule(lr){4-5}
		& w/o ndiv & w ndiv & w/o ndiv & w ndiv \\
		\midrule
		GAN+SN & 23.7 & \textbf{22.9} & 10.5 & \textbf{10.2} \\  
		GAN+SN+GP & 22.9 & \textbf{22.0} & 9.4 & \textbf{9.1} \\
		WGAN+GP & 25.1 & \textbf{23.9} & 9.9 & \textbf{9.5} \\
		WGAN+SN+GP & 23.7 & \textbf{23.3} & 9.2 & \textbf{9.0} \\
		\bottomrule
	\end{tabular}
	\vspace{-10pt}
\end{table}

We conducted experiments on multiple vision tasks under both unsupervised setting (Section \ref{sec::exp_synthetic} and \ref{sec::exp_img_generation}) and conditional setting (Section \ref{sec::exp_text2image} and \ref{sec::handpose}) to demonstrate the effectiveness of the proposed idea. 

Compared to the baseline, we added the normalized diversification loss with all other settings unchanged. We simply used $\alpha=0.5$ for Eq.(\ref{eq::div}) in most experiments except conditional image generation, where we used $\alpha=0.8$.

\subsection{Synthetic Datasets}
\label{sec::exp_synthetic}

We tested our method on the widely used 2D Gaussian ring and 2D Gaussian grid to study the behavior of mode collapse. For each iteration in our method, we performed line 16 and 18 in Algorithm \ref{alg::pipeline_cross} all repeatedly for 3 times for faster convergence. Results are shown in Table \ref{tab::5x5}. We compared several baselines \cite{goodfellow2014generative,metz2016unrolled,srivastava2017veegan,lin2017pacgan,xiao2018bourgan}. Of those methods VEEGAN \cite{srivastava2017veegan}, PacGAN \cite{lin2017pacgan} and BourGAN \cite{xiao2018bourgan} require pairwise information of the real image sets, which is an expensive pre-requisite and cannot fit in the conditional setting. Moreover, BourGAN \cite{xiao2018bourgan} suffered from severely overfitting the training samples (See Fig. \ref{fig::donut-demo}). Our method, with the normalized diversification, achieved promising results in the long-term training (50k iterations) with only on-demand pairwise information. In Fig. \ref{fig::5x5}, we qualitatively demonstrate the results on synthetic point sets where our method achieve comparable or even better results over VEEGAN \cite{srivastava2017veegan} and PacGAN \cite{lin2017pacgan} without using two or more real samples at a time. 

\subsection{Image Generation}
\label{sec::exp_img_generation}


We further tested our method on the widely accepted task of image generation. We conducted experiments on CIFAR-10 \cite{krizhevsky2009learning} and CelebA \cite{liu2015deep} with state-of-the-art methods via the off-the-shelf library\footnote{https://github.com/google/compare\_gan}. Because there exist relatively dense samples in this setting, conventional methods fit the problem relatively well. Nevertheless, as shown in Table \ref{tab::gan_multiple}, with the normalized diversification loss, our method achieves consistent improvement on CIFAR-10 \cite{krizhevsky2009learning} and CelebA \cite{liu2015deep} over strong baselines with spectral normalization \cite{miyato2018spectral} and gradient penalty \cite{gulrajani2017improved}. By regularizing the model behavior around the visited modes through normalized diversification, the generator maintains a safe extrapolation depending on the current outputs.


\subsection{Conditional Image Generation}
\label{sec::exp_text2image}
We tested the performance of conditional image generation described in Section \ref{sec::method_text2image}. For text-to-image translation, we directly followed the architecture used in \cite{reed2016generative} for fair comparison and added the diversity loss term to the overall objective function. We used a discriminator as $h_Y$ to extract features from synthesized images and measured the pairwise image distance in feature space. Experiments were conducted on Oxford-102 Flower dataset \cite{nilsback2008automated} with 5 human-generated captions per image. We generated 5 samples for each text at training stage. Both the baseline and our method were trained for 100 epochs on the training set. We employed the zero-shot setting where we used the 3979 test sets to evaluate both methods. We evaluated the quality and the diversity of the generated images. For the diversity measurement, we first generated 10 images for each input text, then used the perceptual metric \cite{zhangunreasonable} via Inception model \cite{Szegedy_2015_CVPR} to compute the variance of the 10 generated samples and took average over test sets. Table \ref{tab::us_text2image} shows the results on \cite{nilsback2008automated}. With comparable image quality, we significantly improved upon the baseline in terms of the diversity.

\begin{figure}[tb]
\centering
\scriptsize
\setlength\tabcolsep{1pt} 
\begin{tabular}{c|@{\hskip 0.15in}|ccccc}
   {\includegraphics[width=0.15\linewidth, height=2cm]{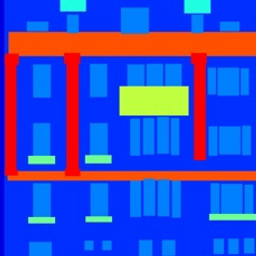}} 
  &
 {\includegraphics[width=0.15\linewidth, height=2cm]{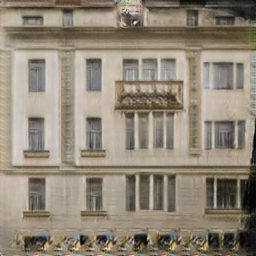}} &
 {\includegraphics[width=0.15\linewidth, height=2cm]{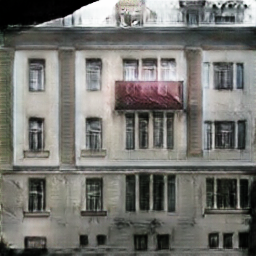}} &
 {\includegraphics[width=0.15\linewidth, height=2cm]{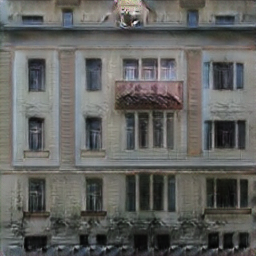}} &
 {\includegraphics[width=0.15\linewidth, height=2cm]{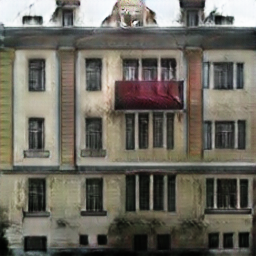}} &
 {\includegraphics[width=0.15\linewidth, height=2cm]{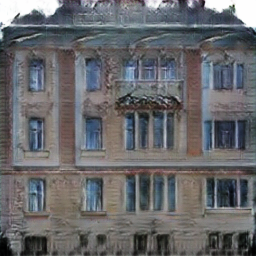}}
 \\
 {\includegraphics[width=0.15\linewidth, height=2cm]{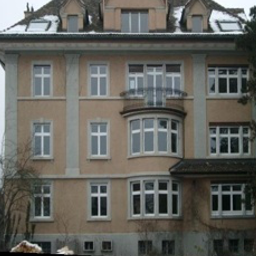}} &
 {\includegraphics[width=0.15\linewidth, height=2cm]{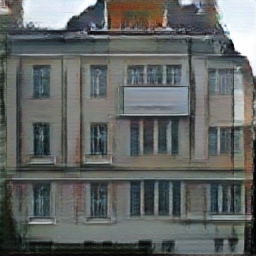}} &
 {\includegraphics[width=0.15\linewidth, height=2cm]{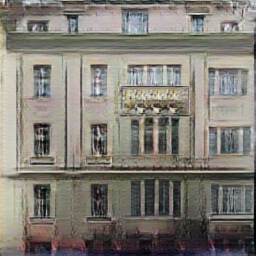}} &
 {\includegraphics[width=0.15\linewidth, height=2cm]{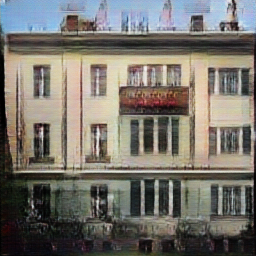}} &
 {\includegraphics[width=0.15\linewidth, height=2cm]{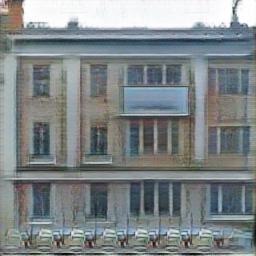}} &
 {\includegraphics[width=0.15\linewidth, height=2cm]{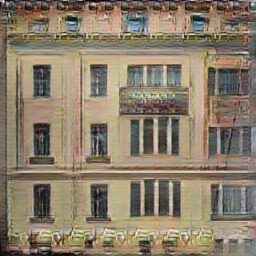}}
  \\
 {\includegraphics[width=0.15\linewidth, height=2cm]{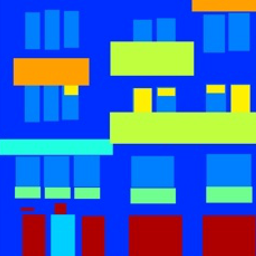}} &
 {\includegraphics[width=0.15\linewidth, height=2cm]{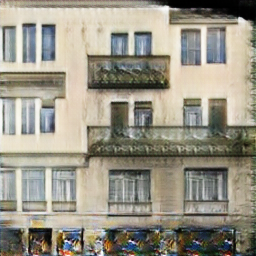}} &
 {\includegraphics[width=0.15\linewidth, height=2cm]{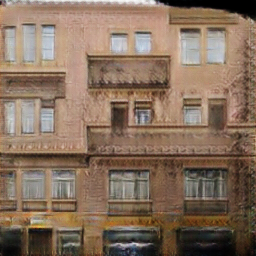}} &
 {\includegraphics[width=0.15\linewidth, height=2cm]{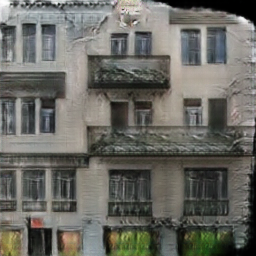}} &
 {\includegraphics[width=0.15\linewidth, height=2cm]{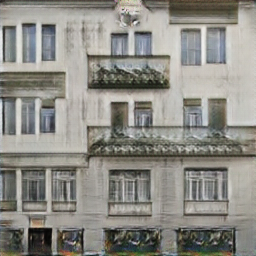}} &
 {\includegraphics[width=0.15\linewidth, height=2cm]{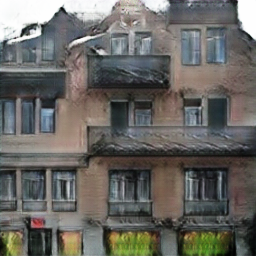}}
 \\
 {\includegraphics[width=0.15\linewidth, height=2cm]{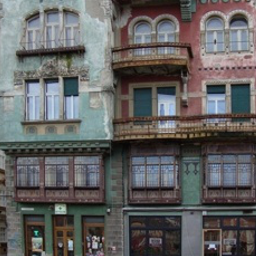}} &
 {\includegraphics[width=0.15\linewidth, height=2cm]{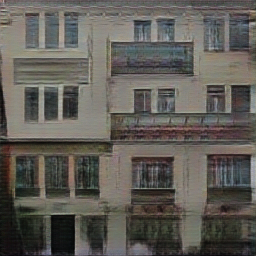}} &
 {\includegraphics[width=0.15\linewidth, height=2cm]{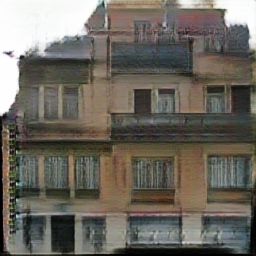}} &
 {\includegraphics[width=0.15\linewidth, height=2cm]{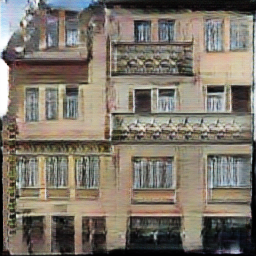}} &
 {\includegraphics[width=0.15\linewidth, height=2cm]{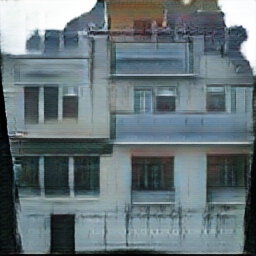}} &
 {\includegraphics[width=0.15\linewidth, height=2cm]{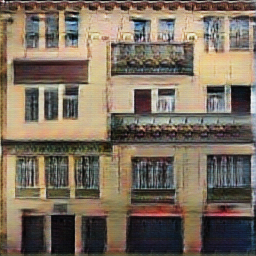}}
\end{tabular}
\centering
\caption{Qualitative results of conditional image generation on facade dataset \cite{cordts2016cityscapes}. Our method improves both image quality and diversity over BicycleGAN \cite{zhu2017toward}. \textbf{Top left:} input image. \textbf{Bottom Left:} corresponding groundtruth image. \textbf{Top Right:} five generated images from BicycleGAN \cite{zhu2017toward}. \textbf{Bottom Right:} five generated images from Ours.}
\label{fig::image2image}
\vspace{-10pt}
\end{figure}

We compare with a strong baseline, BicycleGAN \cite{zhu2017toward}, on image-to-image translation to further demonstrate our ability to model multimodal distribution. We removed the conditional latent regression branch of BicycleGAN \cite{zhu2017toward} and added normalized diversification. We used $\ell_1$ as the pairwise image distance. Table \ref{tab::image2image} and Fig. \ref{fig::image2image} show the results on \cite{cordts2016cityscapes}. Even without the conditional latent regression branch \cite{zhu2017toward}, our method outperforms BicycleGAN \cite{zhu2017toward} on both image quality and diversity. 

\begin{table}[tb]
	\centering
	\caption{Results of text-to-image translation on \cite{nilsback2008automated}. 15 volunteers voted for the better of the randomly sampled pairs for a random text on the webpage. Diversity was computed via the perceptual metric \cite{zhangunreasonable} using Inception model \cite{Szegedy_2015_CVPR}.}
	\label{tab::us_text2image}
	\begin{tabular}{|c||c|c|}
		\hline
		& votes  & diversity \\
		\hline
		GAN-CLS \cite{reed2016generative} & 287 (31.3\%) & 43.7$\pm$16.8 \\
		\hline
		Ours & \textbf{291} (\textbf{31.7}\%) & \textbf{58.1}$\pm$18.5 \\
		\hline
		Neutral & 340 (37.0\%) & - \\
		\hline
	\end{tabular}
\end{table}

\begin{table}[tb]
	\centering
	\caption{Quantitative comparison with BicycleGAN \cite{zhu2017toward} on conditional facade image generation \cite{cordts2016cityscapes}. We computed FID \cite{NIPS2017_7240} with the real data to measure the image quality (lower better) and used LPIPS \cite{zhangunreasonable} to measure the output diversity (higher better). }
	\label{tab::image2image}
	\begin{tabular}{|p{80pt}||p{40pt}|p{60pt}|}
		\hline
		& FID $\downarrow$ & LPIPS $\uparrow$\\
		\hline
		Real data & 0.0 & 0.291 \\
		\hline
		BicycleGAN \cite{zhu2017toward} & 83.0 & 0.136 $\pm$ 0.049 \\
		\hline
		Ours & \textbf{76.3} & \textbf{0.188} $\pm$ 0.048 \\
		\hline
	\end{tabular}
	\vspace{-5pt}
\end{table}

\subsection{Hand Pose Estimation}
\begin{figure*}[tb]
	\begin{minipage}[tb]{0.27\textwidth}
		\centering
		\vspace{-20pt}
		\captionsetup{type=table}
		\caption{Visible joint accuracy on GANerated hands \cite{mueller2018ganerated}. Using a conditional model along with the proposed normalized diversification helps on the visible joints predictions. Our method achieves 1.6 (91.6$\rightarrow$93.2) accuracy gain over deterministic regression.}
		\label{tab::ganerated_auc}
		\begin{tabular}{l||c}
			\hline
			Regression & 91.6 \\
			\hline
			Ours w/o diversity loss & 91.7 \\
			\hline
			Ours & 92.8 \\
			\hline
			Ours+ & \textbf{93.2} \\
			\hline 
		\end{tabular}
		\vspace{-20pt}
	\end{minipage} \ \ \ \ 
	\begin{minipage}[tb]{0.34\textwidth}
		\centering
		\vspace{-15pt}
		\includegraphics[width=\textwidth]{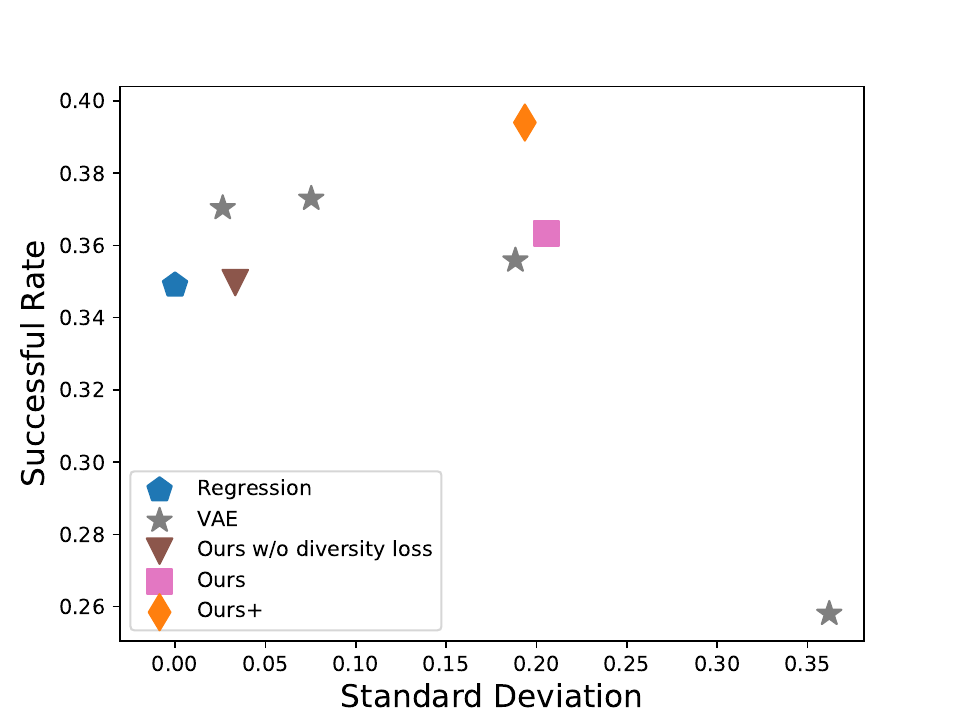}
		\caption{Results on successful rate and standard deviation. We gets significant improvements on both quality and diversity.}
		\label{fig::std_auc}
	\end{minipage} \ \ \ \
	\begin{minipage}[tb]{0.34\textwidth}
		\centering
		\vspace{-15pt}
		\includegraphics[width=\textwidth]{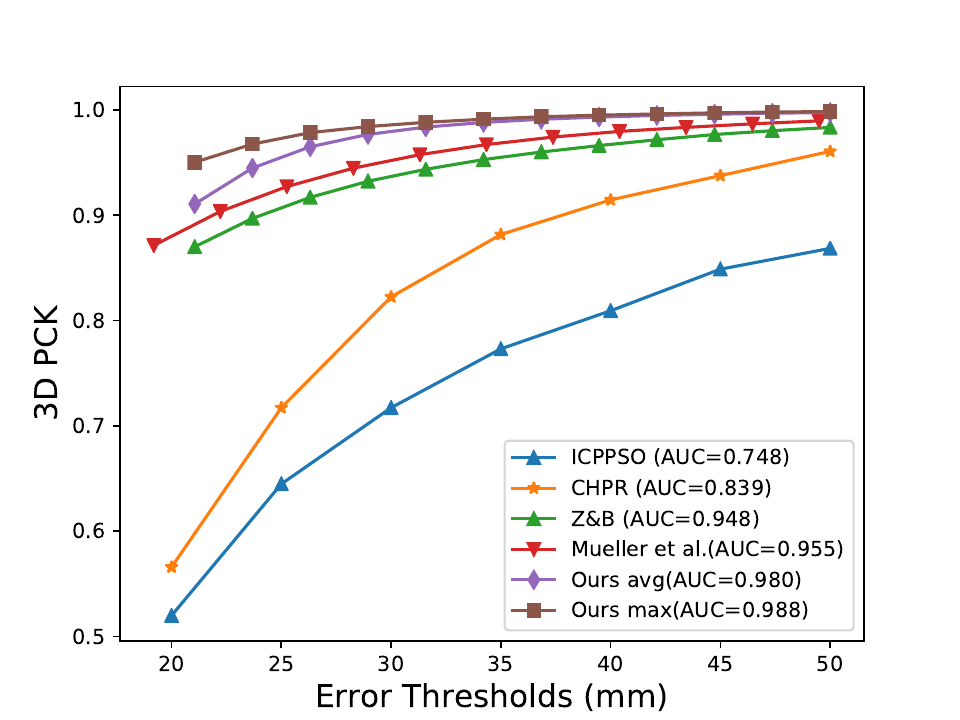}
		\caption{Results on Stereo dataset \cite{zhang20163d}. Our method outperforms existing state-of-the-art alternatives by a large margin.}
		\label{fig::stereo}
	\end{minipage}
\end{figure*}

\begin{figure*}[tb]
	\centering
	\includegraphics[width=500pt,trim={0 0 0 0},clip]{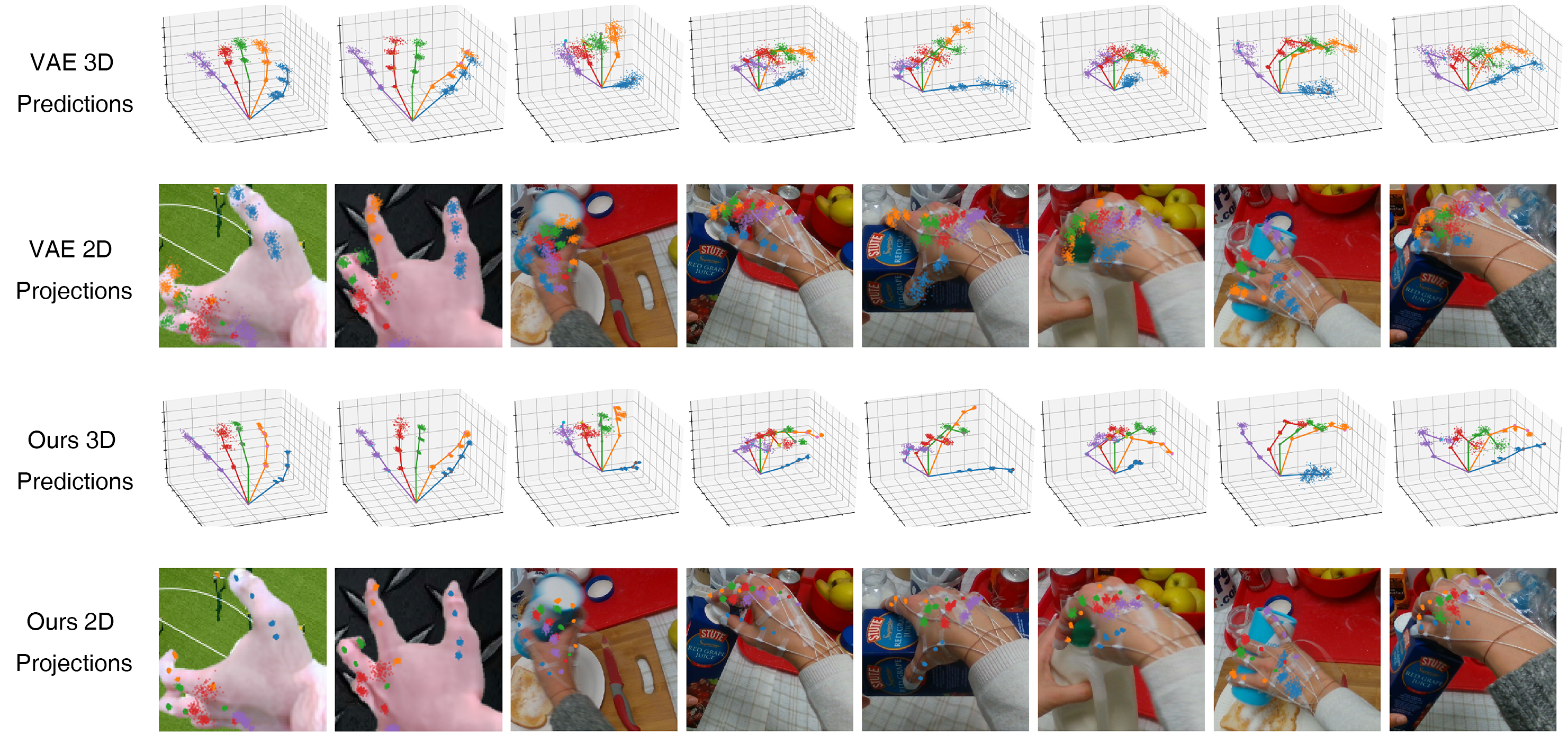}
	\caption{Qualitative results of the multiple pose predictions on GANerated hands \cite{mueller2018ganerated} and FPHAB \cite{garcia2017firstperson}. We show 3D hand predictions and its projections on 2D image (better viewed when zoomed in). With comparable variance, while VAE \cite{kingma2013auto} failed to generated high quality samples, our model generates multiple valid 3D poses with subtle image-pose correlations. }
	\label{fig::hand_multimodal}
	\vspace{-5pt}
\end{figure*}

\label{sec::handpose}
We conducted experiments on three RGB hand datasets including GANerated hands \cite{mueller2018ganerated}, Stereo \cite{zhang20163d}, and FPHAB \cite{garcia2017firstperson}. 
For all datasets, we manually cropped the padded hand bounding box and resized the input to 128x128. Following \cite{mueller2018ganerated}, we directly used the net architecture of their released model. The weights were initialized from ImageNet pretrained model. For each image, our multimodal system generated 20 samples for training and 100 samples for testing. We used $z_{dim}=10$ for all the conditional model. Our method used channel-wise concatenation to aggregate the encoded features $c_i$ and $z_i$. Each dimension of $z_i$ was sampled from a uniform distribution $U(0,1)$. 

Evaluating the multimodal predictions is a non-trivial task. Conventional methods put a max operation on top of the multiple predictions and use the samples nearest the groundtruth for evaluation. Some works even use this max operation over each joint. This results in relatively unfair comparison because simply drawing a sample uniformly distributed in $U(0,1)$ will result in near zero error. Thus, we introduce a better evaluation protocol for the multimodal hand pose estimation: {1) Visible joint accuracy.} For the visible joints, we use the Percentage of Correct Keypoints (PCK) following conventional methods. {2) Standard deviation.} We compute standard deviation of the outputs for each image and take average. {3) Successful rate.} For each image, we use the pre-computed hand mask and object mask to check validity of the samples. The sample is considered valid if the whole hand configuration including the occluded joints lie inside the foreground mask. 

We first compare our method with the deterministic regression method on the GANerated hand dataset \cite{mueller2018ganerated}. 
We concatenate the encoded feature $c_i\in \mathbb{R}^{100}$ with $z_i$ at bottleneck which is contrast to the architecture of VAE \cite{kingma2013auto} where they restricted the latent space $z_i$ to be in the form of Gaussian distribution. `Ours+' denotes our model adapting a higher dimensional latent space $c_i\in \mathbb{R}^{16384}$. We use 100 dimensional latent space $z_i$ in all experiments. 
Combining Table \ref{tab::ganerated_auc} and Figure \ref{fig::std_auc}, it is clear that using a one-to-many conditional model benefits much on the visible joints as well as the successful rate. Our method achieves significant improvements on all three metrics, while VAE \cite{kingma2013auto} struggles more on its quality-diversity trade-off. 

Moreover, we tested our method on two real-world datasets including the Stereo Tracking Benchmark \cite{zhang20163d} and FPHAB \cite{garcia2017firstperson}. Our method outperforms state-of-the-art methods both quantitatively and qualitatively. As is shown in Figure \ref{fig::hand_multimodal}, our method captures much subtle ambiguity and could generate accurate predictions for visible joints on each sample. With normalized diversification, the model maintains a centralized structure which has good property for pairwise interpolation. This promotes ``safer'' extrapolation for robust occluded joints detection which leads to valid yet diversified outputs.

\section{Conclusion}

In this paper, we proposed \textit{normalized diversification}, a generalized loss on the mapping function measuring whether the mapping preserves the relative pairwise distance to address the problem of mode collapse. We aim to diversify the outputs with normalized pairwise distance, encouraging safe interpolation in the latent space and active extrapolation towards outer important states simultaneously. Results show that by employing different metric spaces, normalized diversification can be applied to multiple vision applications and achieves consistent improvements on both encoding quality and output diversity. 


\subsection*{Acknowledgements}

We gratefully appreciate the support from Honda Research Institute Curious Minded Machine Program. We sincerely thank Dr. Gedas Bertasius and Xin Yuan for valuable discussions.

{\small
	\bibliographystyle{ieee}
	\bibliography{diversity}

\begin{thebibliography}{10}\itemsep=-1pt

\bibitem{arjovsky2017wasserstein}
M.~Arjovsky, S.~Chintala, and L.~Bottou.
\newblock Wasserstein generative adversarial networks.
\newblock In {\em ICML}, pages 214--223, 2017.

\bibitem{bansal2017pixelnn}
A.~Bansal, Y.~Sheikh, and D.~Ramanan.
\newblock Pixelnn: Example-based image synthesis.
\newblock 2018.

\bibitem{bao2017cvae}
J.~Bao, D.~Chen, F.~Wen, H.~Li, and G.~Hua.
\newblock Cvae-gan: Fine-grained image generation through asymmetric training.
\newblock In {\em ICCV}, pages 2745--2754, 2017.

\bibitem{berthelot2017began}
D.~Berthelot, T.~Schumm, and L.~Metz.
\newblock Began: boundary equilibrium generative adversarial networks.
\newblock {\em arXiv preprint arXiv:1703.10717}, 2017.

\bibitem{che2015mode}
T.~Che, Y.~Li, A.~P. Jacob, Y.~Bengio, and W.~Li.
\newblock Mode regularized generative adversarial networks.
\newblock In {\em ICLR}, 2017.

\bibitem{chen2017photographic}
Q.~Chen and V.~Koltun.
\newblock Photographic image synthesis with cascaded refinement networks.
\newblock In {\em ICCV}, volume~1, page~3, 2017.

\bibitem{cordts2016cityscapes}
M.~Cordts, M.~Omran, S.~Ramos, T.~Rehfeld, M.~Enzweiler, R.~Benenson,
  U.~Franke, S.~Roth, and B.~Schiele.
\newblock The cityscapes dataset for semantic urban scene understanding.
\newblock In {\em CVPR}, pages 3213--3223, 2016.

\bibitem{davis2007information}
J.~V. Davis, B.~Kulis, P.~Jain, S.~Sra, and I.~S. Dhillon.
\newblock Information-theoretic metric learning.
\newblock In {\em ICML}, pages 209--216. ACM, 2007.

\bibitem{durugkar2016generative}
I.~Durugkar, I.~Gemp, and S.~Mahadevan.
\newblock Generative multi-adversarial networks.
\newblock In {\em ICLR}, 2017.

\bibitem{dziugaite2015training}
G.~K. Dziugaite, D.~M. Roy, and Z.~Ghahramani.
\newblock Training generative neural networks via maximum mean discrepancy
  optimization.
\newblock In {\em UAI}, 2015.

\bibitem{fan2017point}
H.~Fan, H.~Su, and L.~J. Guibas.
\newblock A point set generation network for 3d object reconstruction from a
  single image.
\newblock In {\em CVPR}, volume~2, page~6, 2017.

\bibitem{firman2018diversenet}
M.~Firman, N.~D. Campbell, L.~Agapito, and G.~J. Brostow.
\newblock Diversenet: When one right answer is not enough.
\newblock In {\em CVPR}, pages 5598--5607, 2018.

\bibitem{garcia2017firstperson}
G.~Garcia-Hernando, S.~Yuan, S.~Baek, and T.-K. Kim.
\newblock Firstperson hand action benchmark with rgb-d videos and 3d hand pose
  annotations.
\newblock In {\em CVPR}, 2018.

\bibitem{goodfellow2014generative}
I.~Goodfellow, J.~Pouget-Abadie, M.~Mirza, B.~Xu, D.~Warde-Farley, S.~Ozair,
  A.~Courville, and Y.~Bengio.
\newblock Generative adversarial nets.
\newblock In {\em NIPS}, pages 2672--2680, 2014.

\bibitem{gulrajani2017improved}
I.~Gulrajani, F.~Ahmed, M.~Arjovsky, V.~Dumoulin, and A.~C. Courville.
\newblock Improved training of wasserstein gans.
\newblock In {\em NIPS}, pages 5767--5777, 2017.

\bibitem{NIPS2017_7240}
M.~Heusel, H.~Ramsauer, T.~Unterthiner, B.~Nessler, and S.~Hochreiter.
\newblock Gans trained by a two time-scale update rule converge to a local nash
  equilibrium.
\newblock In {\em NIPS}, pages 6629--6640. 2017.

\bibitem{huang2018multimodal}
X.~Huang, M.-Y. Liu, S.~Belongie, and J.~Kautz.
\newblock Multimodal unsupervised image-to-image translation.
\newblock In {\em ECCV}, 2018.

\bibitem{isola2017image}
P.~Isola, J.-Y. Zhu, T.~Zhou, and A.~A. Efros.
\newblock Image-to-image translation with conditional adversarial networks.
\newblock In {\em CVPR}, 2017.

\bibitem{STN}
M.~Jaderberg, K.~Simonyan, A.~Zisserman, et~al.
\newblock Spatial transformer networks.
\newblock In {\em NIPS}, pages 2017--2025, 2015.

\bibitem{johnson2016perceptual}
J.~Johnson, A.~Alahi, and L.~Fei-Fei.
\newblock Perceptual losses for real-time style transfer and super-resolution.
\newblock In {\em European conference on computer vision}, pages 694--711.
  Springer, 2016.

\bibitem{karras2017progressive}
T.~Karras, T.~Aila, S.~Laine, and J.~Lehtinen.
\newblock Progressive growing of gans for improved quality, stability, and
  variation.
\newblock In {\em ICLR}, 2018.

\bibitem{kingma2013auto}
D.~P. Kingma and M.~Welling.
\newblock Auto-encoding variational bayes.
\newblock {\em arXiv preprint arXiv:1312.6114}, 2013.

\bibitem{krizhevsky2009learning}
A.~Krizhevsky and G.~Hinton.
\newblock Learning multiple layers of features from tiny images.
\newblock 2009.

\bibitem{larsen2016autoencoding}
A.~B.~L. Larsen, S.~K. S{\o}nderby, H.~Larochelle, and O.~Winther.
\newblock Autoencoding beyond pixels using a learned similarity metric.
\newblock In {\em ICML}, pages 1558--1566, 2016.

\bibitem{lin2014microsoft}
T.-Y. Lin, M.~Maire, S.~Belongie, J.~Hays, P.~Perona, D.~Ramanan,
  P.~Doll{\'a}r, and C.~L. Zitnick.
\newblock Microsoft coco: Common objects in context.
\newblock In {\em ECCV}, pages 740--755. Springer, 2014.

\bibitem{lin2017pacgan}
Z.~Lin, A.~Khetan, G.~Fanti, and S.~Oh.
\newblock Pacgan: The power of two samples in generative adversarial networks.
\newblock In {\em NeurIPS}, 2018.

\bibitem{liu2015deep}
Z.~Liu, P.~Luo, X.~Wang, and X.~Tang.
\newblock Deep learning face attributes in the wild.
\newblock In {\em ICCV}, pages 3730--3738, 2015.

\bibitem{magurran1988diversity}
A.~E. Magurran.
\newblock Why diversity?
\newblock In {\em Ecological diversity and its measurement}, pages 1--5.
  Springer, 1988.

\bibitem{metz2016unrolled}
L.~Metz, B.~Poole, D.~Pfau, and J.~Sohl-Dickstein.
\newblock Unrolled generative adversarial networks.
\newblock In {\em ICLR}, 2017.

\bibitem{miyato2018spectral}
T.~Miyato, T.~Kataoka, M.~Koyama, and Y.~Yoshida.
\newblock Spectral normalization for generative adversarial networks.
\newblock In {\em ICLR}, 2018.

\bibitem{mueller2018ganerated}
F.~Mueller, F.~Bernard, O.~Sotnychenko, D.~Mehta, S.~Sridhar, D.~Casas, and
  C.~Theobalt.
\newblock Ganerated hands for real-time 3d hand tracking from monocular rgb.
\newblock In {\em CVPR}, pages 49--59, 2018.

\bibitem{mueller2017real}
F.~Mueller, D.~Mehta, O.~Sotnychenko, S.~Sridhar, D.~Casas, and C.~Theobalt.
\newblock Real-time hand tracking under occlusion from an egocentric rgb-d
  sensor.
\newblock In {\em ICCV}, volume~10, 2017.

\bibitem{nilsback2008automated}
M.-E. Nilsback and A.~Zisserman.
\newblock Automated flower classification over a large number of classes.
\newblock In {\em Computer Vision, Graphics \& Image Processing, 2008.
  ICVGIP'08. Sixth Indian Conference on}, pages 722--729. IEEE, 2008.

\bibitem{radford2015unsupervised}
A.~Radford, L.~Metz, and S.~Chintala.
\newblock Unsupervised representation learning with deep convolutional
  generative adversarial networks.
\newblock {\em arXiv preprint arXiv:1511.06434}, 2015.

\bibitem{reed2016generative}
S.~Reed, Z.~Akata, X.~Yan, L.~Logeswaran, B.~Schiele, and H.~Lee.
\newblock Generative adversarial text to image synthesis.
\newblock In {\em ICML}, pages 1060--1069, 2016.

\bibitem{NIPS2016_6125}
T.~Salimans, I.~Goodfellow, W.~Zaremba, V.~Cheung, A.~Radford, X.~Chen, and
  X.~Chen.
\newblock Improved techniques for training gans.
\newblock In {\em NIPS}, pages 2234--2242. 2016.

\bibitem{solnik1974not}
B.~H. Solnik.
\newblock Why not diversify internationally rather than domestically?
\newblock {\em Financial analysts journal}, pages 48--54, 1974.

\bibitem{spurr2018cross}
A.~Spurr, J.~Song, S.~Park, and O.~Hilliges.
\newblock Cross-modal deep variational hand pose estimation.
\newblock In {\em CVPR}, pages 89--98, 2018.

\bibitem{srivastava2017veegan}
A.~Srivastava, L.~Valkoz, C.~Russell, M.~U. Gutmann, and C.~Sutton.
\newblock Veegan: Reducing mode collapse in gans using implicit variational
  learning.
\newblock In {\em NIPS}, pages 3308--3318, 2017.

\bibitem{Szegedy_2015_CVPR}
C.~Szegedy, W.~Liu, Y.~Jia, P.~Sermanet, S.~Reed, D.~Anguelov, D.~Erhan,
  V.~Vanhoucke, and A.~Rabinovich.
\newblock Going deeper with convolutions.
\newblock In {\em CVPR}, 2015.

\bibitem{CPM}
S.-E. Wei, V.~Ramakrishna, T.~Kanade, and Y.~Sheikh.
\newblock Convolutional pose machines.
\newblock In {\em CVPR}, pages 4724--4732, 2016.

\bibitem{weinberger2006distance}
K.~Q. Weinberger, J.~Blitzer, and L.~K. Saul.
\newblock Distance metric learning for large margin nearest neighbor
  classification.
\newblock In {\em NIPS}, pages 1473--1480, 2006.

\bibitem{weinberger2006unsupervised}
K.~Q. Weinberger and L.~K. Saul.
\newblock Unsupervised learning of image manifolds by semidefinite programming.
\newblock {\em IJCV}, 70(1):77--90, 2006.

\bibitem{xiao2018bourgan}
C.~Xiao, P.~Zhong, and C.~Zheng.
\newblock Bourgan: Generative networks with metric embeddings.
\newblock In {\em NeurIPS}, 2018.

\bibitem{xing2003distance}
E.~P. Xing, M.~I. Jordan, S.~J. Russell, and A.~Y. Ng.
\newblock Distance metric learning with application to clustering with
  side-information.
\newblock In {\em NIPS}, pages 521--528, 2003.

\bibitem{ye2017occlusion}
Q.~Ye and T.-K. Kim.
\newblock Occlusion-aware hand pose estimation using hierarchical mixture
  density network.
\newblock In {\em ECCV}, 2018.

\bibitem{zhang20163d}
J.~Zhang, J.~Jiao, M.~Chen, L.~Qu, X.~Xu, and Q.~Yang.
\newblock 3d hand pose tracking and estimation using stereo matching.
\newblock {\em arXiv preprint arXiv:1610.07214}, 2016.

\bibitem{zhangunreasonable}
R.~Zhang, P.~Isola, A.~A. Efros, E.~Shechtman, and O.~Wang.
\newblock The unreasonable effectiveness of deep features as a perceptual
  metric.
\newblock In {\em CVPR}, 2018.

\bibitem{zhu2017toward}
J.-Y. Zhu, R.~Zhang, D.~Pathak, T.~Darrell, A.~A. Efros, O.~Wang, and
  E.~Shechtman.
\newblock Toward multimodal image-to-image translation.
\newblock In {\em NIPS}, pages 465--476, 2017.

\bibitem{ziegler2005improving}
C.-N. Ziegler, S.~M. McNee, J.~A. Konstan, and G.~Lausen.
\newblock Improving recommendation lists through topic diversification.
\newblock In {\em WWW}, pages 22--32. ACM, 2005.

\bibitem{zimmermann2017learning}
C.~Zimmermann and T.~Brox.
\newblock Learning to estimate 3d hand pose from single rgb images.
\newblock In {\em ICCV}, pages 4913--4921. IEEE, 2017.

\end{thebibliography}
}

\clearpage
\section*{Appendix}
\appendix
\section{More Implementation Details}
\textbf{Code will be made available.}

\subsection{Synthetic datasets}
For fair comparison, we directly used the experimental setting and the evaluation protocol in \cite{xiao2018bourgan}. The 2D ring has 8 Gaussian grid (std=0.05) equally distributed on a circle with $r=2$, while the 2D grid has 5x5 Gaussian grid (std=0.1) equally distributed in a 8x8 square.

\subsection{RGB Hand datasets}
Our experiments were conducted on three datasets. First, we augmented a large-scale synthetic hand dataset. Then, we tested our method on two real datasets.

\paragraph{GANerated Hands \cite{mueller2018ganerated}.} This dataset is a synthetic dataset with 332k RGB images for hands. Those images were translated from SynthHands \cite{mueller2017real} via cycle consistency \cite{isola2017image}. We collected frequently interacted objects from COCO \cite{lin2014microsoft} and inserted it onto the images without objects to form a new set of with-object data. In this way, we could get the object masks with visibility annotation. Note that currently there is no dataset with visibility annotation available (\cite{ye2017occlusion} does not release their used dataset). For the experiments, we equally (50/50) split the train/test set. The final dataset consists of 143k images for training.

\paragraph{Stereo Hand Benchmark \cite{zhang20163d}.} 
This dataset consists of 12 sequences including 36k rgb images without hand-object interactions. For the experiments, we used the conventional split in \cite{zimmermann2017learning} for direct comparison, where 10 sequences with 30k images were used for training and the rest were used for testing. 

\paragraph{First-person Hand Action Benchmark (FPHAB) \cite{garcia2017firstperson}.}
This large-scale dataset has 1200 sequences. We used the 280 sequences on hand-object interaction with 6-DOF object annotations. Specifically, we used 227 sequences with 17k images for training and 53 sequences with 4k images for testing.

\subsection{Network architecture}
\begin{figure*}
	\centering
	\includegraphics[width=5.7in]{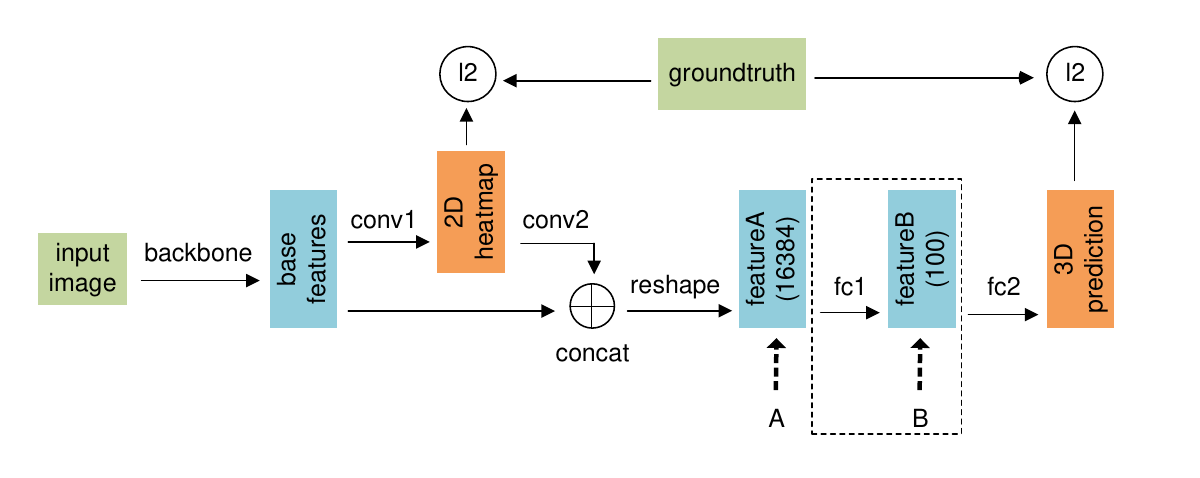}
	\caption{Network architecture for hand pose estimation. For the \textbf{backbone}, we directly borrow the architecture from \cite{mueller2018ganerated}. Following the design protocol of \cite{mueller2018ganerated,CPM}, we use the extracted \textbf{base features} to first reconstruct the \textbf{2D heatmap}. Then the predicted heatmap is concatenated with the base features. `A' and `B' denote where the noise vector $z\in \mathbb{R}^{10}$ is included via concatenation in `Ours+' and `Ours' respectively. The part in the dashed line, which is a bottleneck structure, is not used in `Ours+'. Only visible joints contribute to the $l_2$ distance for both the 2D heatmap and 3D predictions. Specifically, the input image is sized 128x128. `conv1' denotes one stride-1 conv layer and two stride-2 deconv layers. `conv2' denotes two stride-2 conv layers. Both `fc1' and `fc2' denotes two sequential fc layers. The final 3D predictions has a dimension of 21x3=63. }
	\label{fig::arch}
\end{figure*}

For the experiments on synthetic Gaussian datasets, image generation and text-to-image translation, we directly borrow the architecture of the baseline methods \cite{xiao2018bourgan,radford2015unsupervised,reed2016generative} respectively for fair comparison. For the task of hand pose estimation, our network architecture is illustrated in Figure \ref{fig::arch}. We used the same design protocol as \cite{mueller2018ganerated, CPM}, where 2D heatmap was first estimated to guide the 3D joint predictions. When computing the $l_2$ distance, only visible joints were considered. 

\subsection{Differentiable 2D projection}
We used the projected 2D heatmap in the image-pose GAN formulation. However, it is worth noting that simply projecting and transferring the predicted 3D joints into 2D heatmap is non-differentiable. To get the meaningful gradient, we employed the differentiable image sampling technique in \cite{STN}. In this way, we could reparametrize the 2D heatmap with respect to the predicted 3D pose.

\subsection{More detailed experimental settings}


\paragraph{Image-to-image translation.}
{Our implementation is mainly based on the official code\footnote{https://github.com/junyanz/BicycleGAN} provided by \cite{zhu2017toward} where we adapted the same architectures for the generator and the discriminator. Compared to the original BicycleGAN \cite{zhu2017toward} implementation, we removed the image encoder which maps the RGB images to a coding space for re-parametrizing the Gaussian distribution and replaced the instance normalization with spectral normalization \cite{miyato2018spectral}. We followed the same train/val/test split setting as \cite{zhu2017toward}. We trained our model with a batch size of 8 for 325 epochs. We used $\alpha = 0.8$ for normalized diversity loss and 2.0 as L1 reconstruction weighted factor. During training, we randomly sampled 6 codes from the latent space to computed the diversity loss. Our initial learning rate was 2e-4 which was decayed by 0.1 for every 200 epochs. For quantitative evaluation, we randomly selected 100 images from validation set and sampled 38 outputs for each input image. }

\section{More Qualitative Results}
\begin{figure*}
	\centering
	\includegraphics[width=\textwidth]{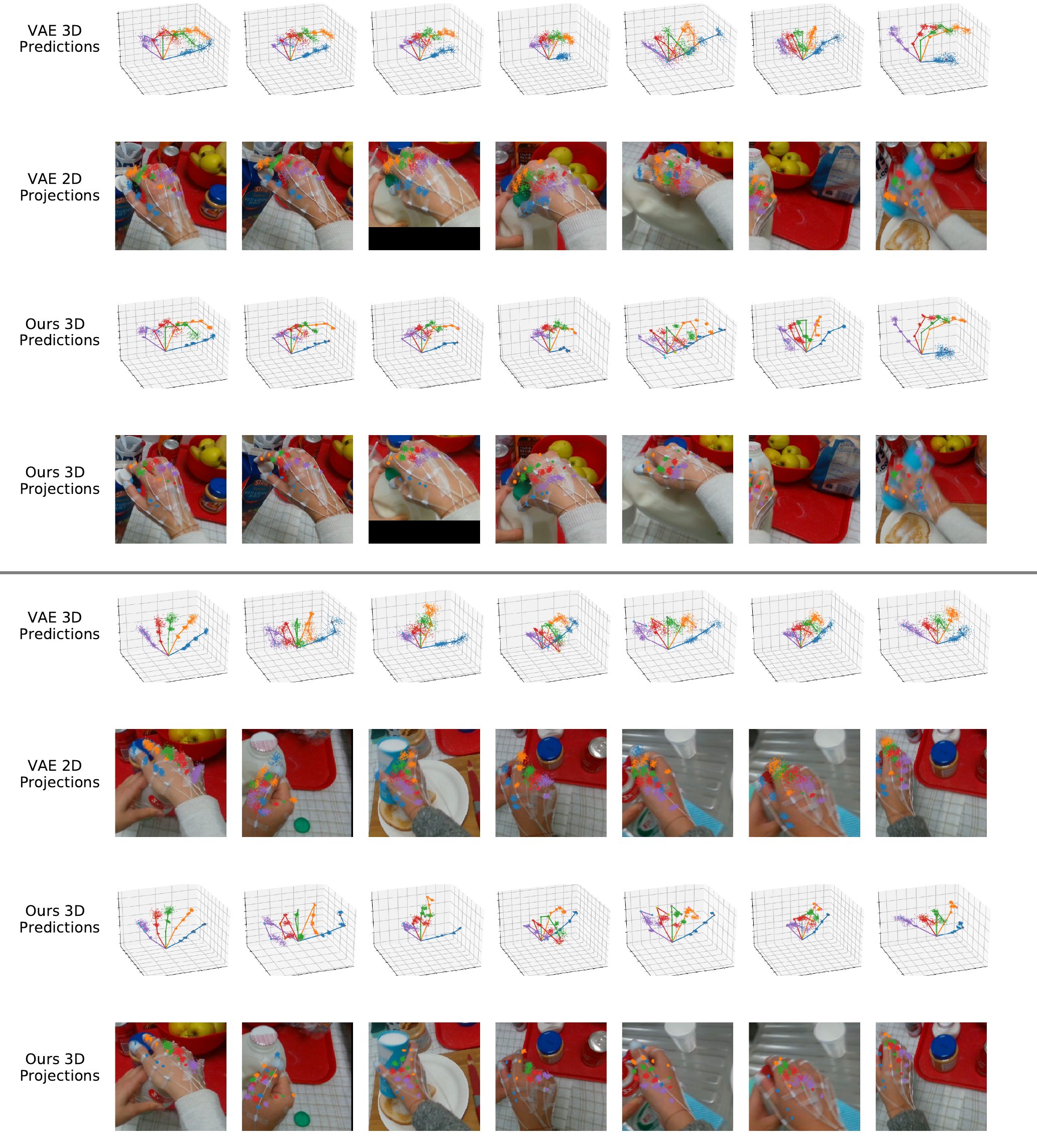}
	\caption{More qualitative comparison between VAE \cite{kingma2013auto} and our method on 3D hand predictions and its projections on 2D image (better viewed when zoomed in).}
	\label{fig::vis_supp}
\end{figure*}

We show more qualitative results on multimodal hand pose estimation from RGB images in Figure \ref{fig::vis_supp}.


\end{document}